%% file: main.tex
\pdfoutput=1

\documentclass[11pt]{article}

\usepackage{EMNLP2023} 

\usepackage{times}
\usepackage{latexsym}

\usepackage[T1]{fontenc}

\usepackage[utf8]{inputenc}

\usepackage{microtype}

\usepackage{inconsolata}
\usepackage{hyperref}

\usepackage{graphicx}
\usepackage{amsthm,amssymb,amsmath}
\usepackage{booktabs}
\usepackage{xltabular}

\usepackage{mathtools}  
\usepackage{amsfonts}  
\usepackage{mathtools}  
\usepackage{amssymb}  

\usepackage{amsmath,amssymb}
\newcommand{\mycomment}[1]{}
\title{Revisiting Instruction Fine-tuned Model Evaluation to Guide Industrial Applications}

\author{Manuel Faysse$^{1,3}$ \quad Gautier Viaud$^{1}$ \quad Céline Hudelot$^{3}$ \quad Pierre Colombo$^{2,3}$ \\ $^{1}$Illuin Technology, Paris, France \quad $^2$Equall, Paris, France \\
$^3$MICS, CentraleSupélec, Université Paris-Saclay, France \,\
\\
\small \url{manuel.faysse@centralesupelec.fr}}

\begin{document}
\maketitle
\begin{abstract}

Instruction Fine-Tuning (IFT) is a powerful paradigm that strengthens the zero-shot capabilities of Large Language Models (LLMs), but in doing so induces new evaluation metric requirements. We show LLM-based metrics to be well adapted to these requirements, and leverage them to conduct an investigation of task-specialization strategies, quantifying the trade-offs that emerge in practical industrial settings. Our findings offer practitioners actionable insights for real-world IFT model deployment.

\end{abstract}

\section{Introduction}
\label{sec:introduction}

Adapting pre-trained language models (LMs) for specific applications is central in industrial NLP to unlock task-specific performance gains and strengthen model alignment with industry requirements.  A paradigm gaining traction is the use of instruction fine-tuned (IFT) models, LMs capable of following arbitrary instructions expressed in natural language \cite{wei2022finetuned, sanh2022multitask, ouyang2022training}.

Researchers primarily concentrate on improving general-purpose IFT models to be used as versatile agents capable of executing instructions expressed in natural language \cite{li2023camel, zhou2023lima, xu2023wizardlm}. 
In an industrial setting, prompting ChatGPT to improve the wording of an email, or to assist with a code snippet would be instances of this zero-shot utilization scenario, which we define as $\mathcal{S}_0$. 
Critical industrial LLM applications may however not always align with $\mathcal{S}_0$, and often prioritize two other settings. The first scenario, $\mathcal{S}_1$, requires extending a generalist IFT model's capabilities to new specific tasks not included in the original instruction training set. The second scenario, $\mathcal{S}_2$, centers around converting IFT models into specialized models proficient \emph{exclusively} on specific tasks.  
%
In $\mathcal{S}_1$ for instance, a large company may want an LLM assistant for internal employee use, and decide to extend an openly available Chat model by training it to write memos with a specific templating scheme, to respond to internal FAQs, and to use internal coding tools, all the while retaining the original chat assistant's general purpose abilities. In $\mathcal{S}_2$, that same company is only interested in a given specific task; extracting specific information from business documents, and specializes an IFT model for that purpose, aiming to leverage prompting and the generalization capabilities of the model for a more data-efficient training.

In this paper, we thoroughly examine $\mathcal{S}_1$ and $\mathcal{S}_2$ by investigating the learning dynamics of specializing IFT models through a practical lens. To ensure the reliability of our tooling and the rigor of our conclusions, we first undertake a critical assessment of the current evaluation practices employed for IFT models. Formally, our contributions are:

\noindent\textbf{Contribution 1.} IFT models are designed to handle tasks of diverse natures and varying difficulties. However, current metrics used to measure their performance are often task-specific \cite{zellers2019hellaswag, eval-harness}, or rely on automatic metrics designed for other intended purposes \cite{papineni-etal-2002-bleu, lin-2004-rouge}. To address this limitation, we introduce two new requirements for metrics used to evaluate IFT models: \underline{C}omparability \underline{A}cross \underline{T}ask (\texttt{CAT}) and \underline{T}ask and \underline{F}ormat \underline{A}gnostism (\texttt{TFA}). \texttt{CAT} imposes for metric scores to exhibit consistency across a diverse set of generative tasks, in contrast to the sole traditional focus of consistency within a specific task. \texttt{TFA} defines the need for metrics to demonstrate robustness to variations in the output formats. By highlighting the shortcomings of existing metrics in meeting \texttt{CAT} and \texttt{TFA}, we present compelling evidence that using LLMs as scoring agents is a viable evaluation alternative of IFT models. 

\noindent\textbf{Contribution 2.} We approach our examination of $\mathcal{S}_1$ and $\mathcal{S}_2$ from a practical perspective and focus on the trade-off between data availability and overall performance. Our analysis uncovers two distinct phases of learning during IFT model specialization: learning to format, and learning to solve tasks. Subsequently, we showcase how practitioners can (i) leverage synthetic data to facilitate learning the desired formatting aspects and (ii) use IFT models to reduce the need of expert data in industrial scenarios. Our study provides practical insights and actionable recommendations to practitioners looking to deploy IFT models in production settings.\footnote{Code and evaluation datasets are available on \url{https://github.com/ManuelFay/IFTEval}.}

\section{Re-evaluating IFT Model Evaluation}

\subsection{What Should Good Scorers Measure?}
\label{sec:evalreq}
In scenarios $\mathcal{S}_0$, $\mathcal{S}_1$, and $\mathcal{S}_2$, IFT models are trained to perform generative tasks. Unlike models designed for single tasks with known output formats, IFT models have the capacity to generate diverse valid responses across different tasks and formats \cite{ouyang2022training}. The novel capabilities of IFT models impose new considerations when selecting an automatic evaluation metric. 
\\\noindent \textbf{Comparability across tasks (\texttt{CAT}).}
Standard evaluation metrics aim to fulfill one key requirement: coherence within each task with respect to human judgment \cite{specia2010machine}. However, due to the multi-task nature of IFT models, the scores should also be comparable across different tasks \cite{colombo2022best,himmi2023towards}. In other words, the scoring scale should be absolute and coherent with human preferences on all tasks. To measure the \texttt{CAT} we will mix samples of different tasks and compute the Spearman correlation ($\rho$) of their score with human judgment\footnote{Common when benchmarking metrics \cite{bhandari-etal-2020-evaluating,colombo2021automatic,chhun2022human,staerman2021pseudo, fabbri-etal-2021-summeval,colombo2021automatic}, we extend the tool to inter-task settings.}. 
This requirement is essential in scenarios $\mathcal{S}_0$ and $\mathcal{S}_1$ to measure model performance across different tasks, and make informed decisions regarding the trade-offs between model variants.
\\\noindent \textbf{Task and Format-Agnostism (\texttt{TFA}).} Evaluation metrics should be robust to artifacts associated with the output format and to the nature of the evaluated task \cite{liang2022holistic}. Implementing task-specific scoring metrics is not a scalable solution for generalist IFT models. To measure \texttt{TFA}, we compute the relative target task improvement between models prompted in a zero-shot manner and models that mastered the task format (trained on 1000 target task samples). Comparing each metric's \texttt{TFA} to human-reported performance improvements allows to grasp the extent to which mastery of the task format influences the metric performance, independently of intrinsic task performance.
 In industrial scenarios, this requirement is essential as it ensures minimal bias in the evaluation due to training data formatting artifacts. In practice, many datasets that industrial actors may add to the training instruction set ($\mathcal{S}_1$), or fully train a model on ($\mathcal{S}_2$) have specific response formatting that differs from what a zero-shot model will answer, leading to a potentially large formatting bias.

\noindent\textbf{Comparability intra-task (\texttt{CIT}).} While in no way a novel requirement, it is essential for metrics to measure performance consistently within a given task.  We verify this by computing the Spearman $\rho$ correlation coefficient between samples of a specific task and human judgments.

In all industrial scenarios for IFT LLMs, rigorous model evaluation is necessarily linked to evaluation metrics that comply with both \texttt{CAT} and \texttt{TFA}, as well as the standard \texttt{CIT} measures.

\subsection{Existing Metrics}
\textbf{Current Evaluation.} Currently, two dominant paradigms emerge for assessing the performance of IFT models: (i) relying on reference-matching scoring metrics such as \texttt{ROUGE-L} \cite{lin-2004-rouge}, or normalized log-probabilities of class labels in few-shot classification benchmarks \cite{hendrycks2021measuring, eval-harness, zellers2019hellaswag}, and (ii) model ranking frameworks, based on pairwise preference comparisons of response quality judged by humans or LLM evaluators \cite{vicuna2023, dubois2023alpacafarm, gudibande2023false}. Language Model based scoring has been shown to be a promising alternative on specific tasks, such as summarization \cite{liu2023geval,colombo2022infolm} or translation \cite{kocmi2023large, xu2023instructscore}. Our work extends these findings to showcase the multi-task scoring capabilities of LLMs with respect to \texttt{CAT} and \texttt{TFA}.

\noindent\textbf{LMs as Viable Scoring Mechanisms.} 
Given the inherently open nature of IFT model generation, we adopt a reference-free setting to ensure unbiased evaluation. We present an input prompt and the corresponding generated response to the LLM\footnote{We rely on LMs available through the OpenAI API (\textit{i.e.}, GPT4 and GPT3.5). See \autoref{sec:metrics} for details.}, prompting it to assign a score on a scale of 0 to 10, subsequently scaling it between 0 and 1 to facilitate comparisons with other evaluation metrics. 

\noindent \textbf{Baseline Metrics.}  We assess the fulfillment of both \texttt{CAT} and \texttt{TFA} by comparing the proposed metrics against well-established \textit{reference-based} metrics, including ROUGE\footnote{ROUGE-1 is used here, it is best on one-word long labels}, BScore \cite{zhang2020bertscore}, and SBERT \cite{reimers2019sentencebert}, as well as a \textit{machine learned} metric, the OpenAssistant Reward Model (RM) \cite{köpf2023openassistant} trained on human preferences. 

\begin{figure*}[ht]
\begin{minipage}{.7\textwidth}
\centering
    \captionof{table}{\emph{(Left)} \textbf{$\rho$ correlation between human scores and metrics} on the summarization task (SUM), on the \emph{human data} tasks individually then averaged in (CIT), and on the concatenated human tasks (CAT) to form inter-task settings. (TFA) denotes relative metric improvement after 1000 target task samples are added to the training set. \emph{(Right)} \textbf{Metric scores} averaged per \emph{synthetic data} category}\vspace{-.3cm}
    \begin{minipage}{.45\textwidth}
      \centering
      \tiny
      \label{tab:main_metrics}
\resizebox{\textwidth}{!}{\begin{tabular}{lrrrr}
    \toprule
     Scorers & SUM & CAT &  CIT & TFA \\ 
    \midrule
    ROUGE & 0.28 & 0.22 & 0.57 & +513.9 \% \\
    BScore & 0.21 & 0.22 & 0.13  & +49.0 \% \\
    SBERT & 0.25 & 0.29 & 0.43 & +86.3 \% \\
    RM & 0.20 & 0.28 & 0.29 & -44\% \\
    \hline
    GPT4 & \textbf{0.45} & \textbf{0.68} & \textbf{0.77} & +2.1 \% \\
    GPT-3.5 & 0.42 & -0.19 & 0.48  & +9.5 \% \\
    \hline
    Human & 0.54 & - & - & +12.0 \% \\
    \bottomrule
    \end{tabular}}
    \end{minipage}\hspace{-.2cm}
    \begin{minipage}{.55\textwidth}
      \centering
        \tiny
            \resizebox{0.92\textwidth}{!}{\begin{tabular}{lrrrrrrr}
    \toprule
    Scorers & Logic & Code  & Memory & Write \\
    \midrule
     ROUGE & \textbf{0.52} & 0.46 & 0.47 & \textit{0.41} \\
     BScore & \textbf{0.73} & \textit{0.71} & \textbf{0.73} & 0.72 \\
     SBERT & 0.80 & 0.74  & \textbf{0.84} & \textit{0.74} \\
     RM & \textbf{0.49} & 0.43 & \textit{0.28} & 0.33 \\
     \hline
     GPT4 & \textit{0.71} & 0.79  & 0.93 & \textbf{0.97} \\
     GPT-3.5 & \textbf{0.88} & \textit{0.82} & 0.87 & 0.86 \\
    \bottomrule
    \end{tabular}}
    \end{minipage} \hfill
\end{minipage}
    \begin{minipage}{.28\textwidth}
    \includegraphics[width=\textwidth]{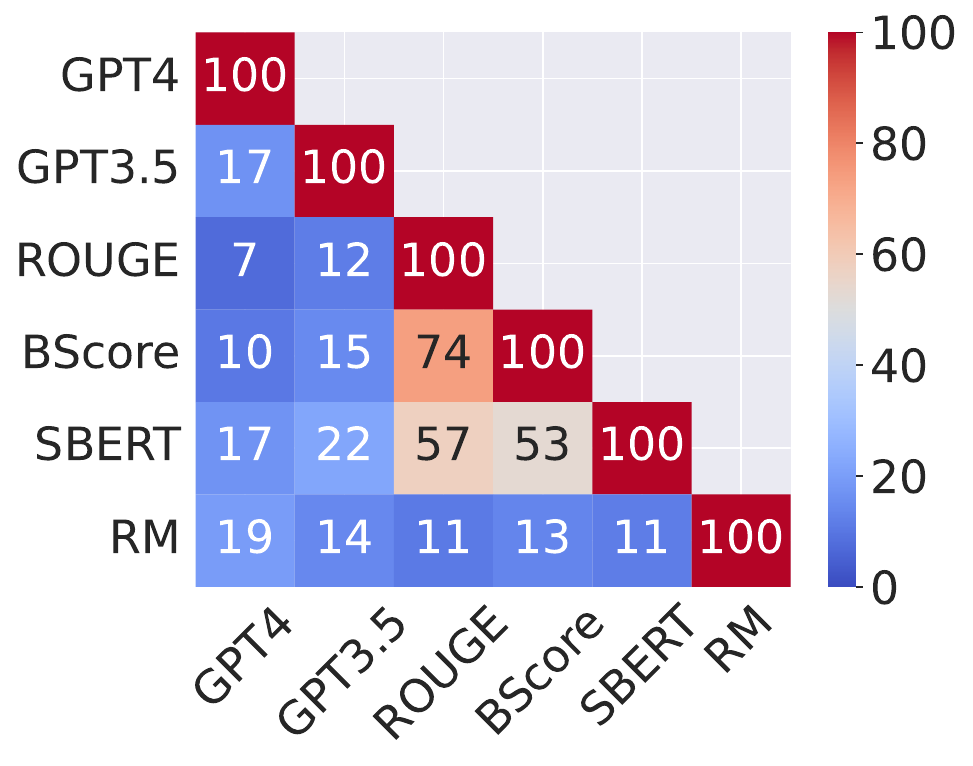}\vspace{-.3cm}
    \captionof{figure}{\textbf{Spearman $\rho$ between metrics} on synthetic data.}
    \label{fig:correlations}
    \end{minipage}
\end{figure*}
\subsection{Experimental setup}

\textbf{Training an IFT model.} IFT models are trained by fine-tuning a base model on a large instruction corpus, collected either through human annotations \cite{ouyang2022training, köpf2023openassistant} or concatenating task-specific datasets \cite{sanh2022multitask, mishra2022crosstask}. In line with recent work \cite{vicuna2023,wang2023selfinstruct,peng2023instruction}, we leverage synthetic data as the base instruction set in our IFT models \cite{alpaca2023}.

\noindent\textbf{Benchmarking automatic metrics.} To benchmark the metrics, we rely on a combination of synthetic and real data.  For \emph{synthetic data}, we use the Alpaca GPT4 dataset \citep{alpaca2023}, and tag the data in 13 task categories (see \autoref{sec:alpaca-pres}) (\textit{e.g.}, logic, code, rewrite).  For \emph{human data}, we focus on tasks with industrial interests. Specifically, we include Natural Language Inference \cite{N18-1101, wang2019glue}, Question Answering \cite{rajpurkar2016squad}, NER \cite{tjong-kim-sang-de-meulder-2003-introduction}, and Sentiment Classification \cite{socher2013recursive, agirre-etal-2013-sem}). To build our metric evaluation dataset, we train and run LLaMA-7B models \cite{touvron2023llama} on varied data mixtures and target tasks. For rigor, we also report scores on the summarization with human feedback dataset from \cite{stiennon2022learning} (\texttt{SUM}).\footnote{More details in \autoref{sec:exp_setup_1})}

\subsection{Experimental results}
To better understand the limitation of existing metrics we conduct both single-task analysis to ensure that metrics are able to score tasks reliably as well as multi-task analysis, which is the standard setting for IFT models. Results are reported in Tab. \ref{tab:main_metrics}.

\noindent\textbf{\texttt{CIT} Analysis.} From Tab. \ref{tab:main_metrics}(left, \texttt{SUM}), we observe that the average correlation with human scores for evaluated summaries are higher for LLM models than with traditionally used metrics. Intra-task correlations on all other \emph{human data} tasks, averaged in \texttt{CIT} lead to similar conclusions.

\noindent\textbf{\texttt{CAT} Analysis.} Tab. \ref{tab:main_metrics}(left) shows that all metrics, \emph{with the exception of the GPT4-based metric}, exhibit weak or no correlation in the context of inter-task consistency. While it is true that existing metrics demonstrate the ability to differentiate between good and bad samples within a single task (\texttt{CIT}), their \emph{performance falls short when confronted with the open setting imposed by IFT models}.

\noindent\textbf{\texttt{TFA} Analysis.} On non-LLM-based metrics, performance gains reported between zero-shot models, and models trained on 1000 target-task samples (Tab. \ref{tab:main_metrics}(left), \texttt{TFA}) largely exceed the  12.0 \% relative improvement of human scores, and demonstrate how format, once learned, unrealistically boosts reference-based metrics which are heavily impacted by format.

\noindent\textbf{Metric Similarity Analysis.} \autoref{fig:correlations} displays metric correlation at the sample level on the synthetic dataset. The results align with \cite{chen2022reproducibility}, indicating a moderate to high correlation between BERT-based metrics and ROUGE-L. However, all metrics exhibit a low correlation with GPT4, indicating different response features are scored. 

\noindent\textbf{Zoom on GPT4.} Tab.\ref{tab:main_metrics}(right) shows a strong correlation between the results of GPT4-based metrics and the corresponding LLM task abilities reported in \citet{wei2022emergent} (Logic and Coding are non-trivial for LLMs, Writing tasks are relatively easier). However, reference-based metrics such as ROUGE suggest the opposite, as they are biased by the high syntactic overlap between model outputs and reference answers in these categories. The GPT3.5 scorer also highly rates results on the Logical Reasoning tasks, contrarily to GPT4. This is due to its lesser ability to spot logical inconsistencies in the evaluated responses \cite{bubeck2023sparks}, hinting that evaluator models must be capable at the evaluated task themselves in order to produce meaningful scores.


Our findings highlight the inadequacy of existing metrics in quantifying the performance of IFT models, while emphasizing GPT4 as a promising candidate. This performance gap in evaluation capabilities is primarily explained by GPT4's reduced dependence to reference answers, leading to a more coherent and absolute evaluation scale \texttt{CAT}, and an improved robustness to variations in output formatting \texttt{TFA}. \emph{The GPT4 scorer's powerful capabilities unlock the study of novel settings traditional metrics would struggle with} \cite{schaeffer2023emergent}.

\section{IFT Models for Industrial Applications}

\subsection{$\mathcal{S}_1$: Improving Specific Tasks}\label{ssec:scenario1}
In this section, we delve into $\mathcal{S}_1$, which specifically aims to extend an IFT model's capabilities to better perform on specific instructions.


\noindent\textbf{Setting.} We fine-tune a base 7B-LLM model (Pythia \citet{biderman2023pythia}, Bloom \citet{scao2022bloom}, Falcon \citet{penedo2023refinedweb}, or LLaMA) using synthetic instructions. In each training iteration, we introduce a selected number of $N$ \emph{real} target task samples into the synthetic dataset. We evaluate the model performance on an independent test subset of the target task.\footnote{More experimental details are given in \autoref{sec:exp_setup_app_31}.}
\begin{figure}[ht]
    \centering
    \includegraphics[width=0.5\textwidth]{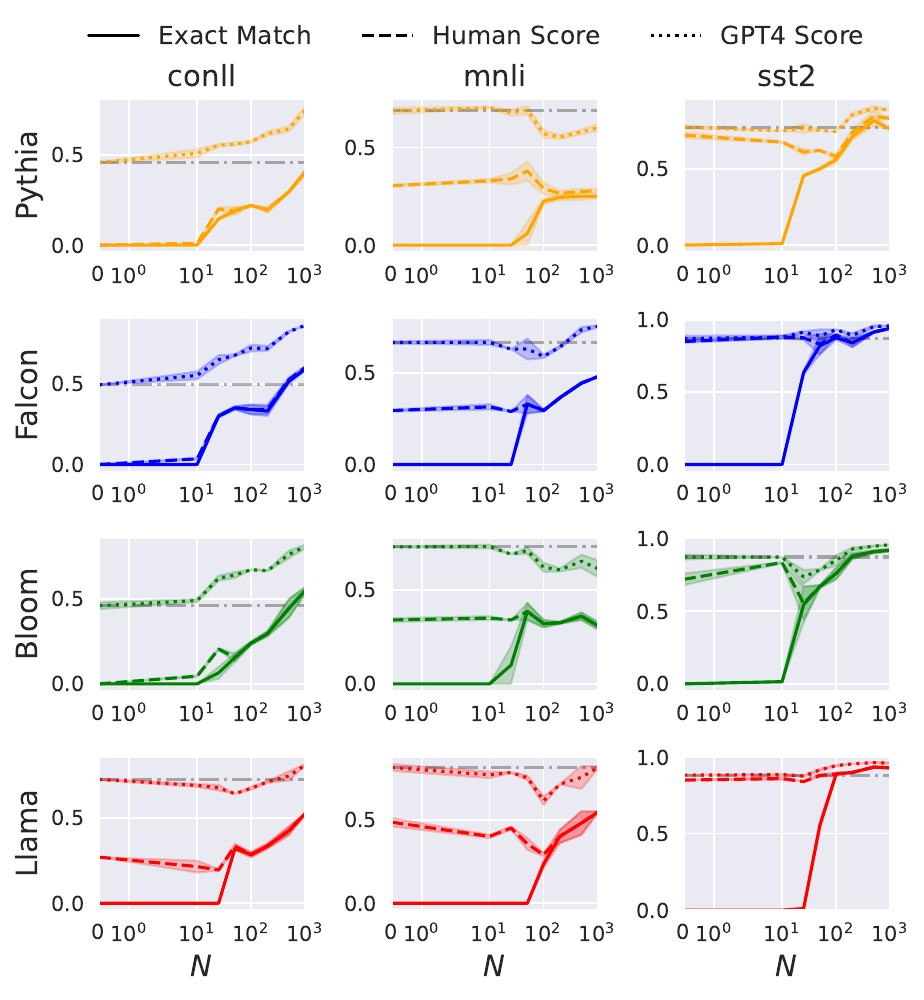}
    \caption{Incorporating $0 \leq N \leq 1000$ real task samples into IFT model training}
    \label{fig:exp1b_all}
\end{figure} 

\noindent\textbf{Mastering Format to Foster Understanding.} \autoref{fig:exp1b_all} shows target task performance as the number of target task samples introduced within the base training set increases. Across all tasks and models, \emph{specialization is biphasic}: first, \textit{task output format} is learned while overall performance remains constant, or even slightly decreases. Only once the format has been mastered, as noted by the spike of the Exact Match, does the model improve upon its \textit{underlying task performance} (Human and GPT4 scores increase), eventually surpassing the original zero-shot performance. It is worth noting that this analysis is made possible by format-agnostic scorers (\texttt{TFA}) that can accurately decouple output format from underlying model performance.

\noindent\textbf{Measuring Model Forgetting.} Performance on a test split of the Alpaca data shows little to no performance degradation (<1\%) caused by the inclusion of new tasks to the training mix (\autoref{sec:app_perf_deg}).

\begin{figure}[ht]
    \centering
    \includegraphics[width=0.45\textwidth]{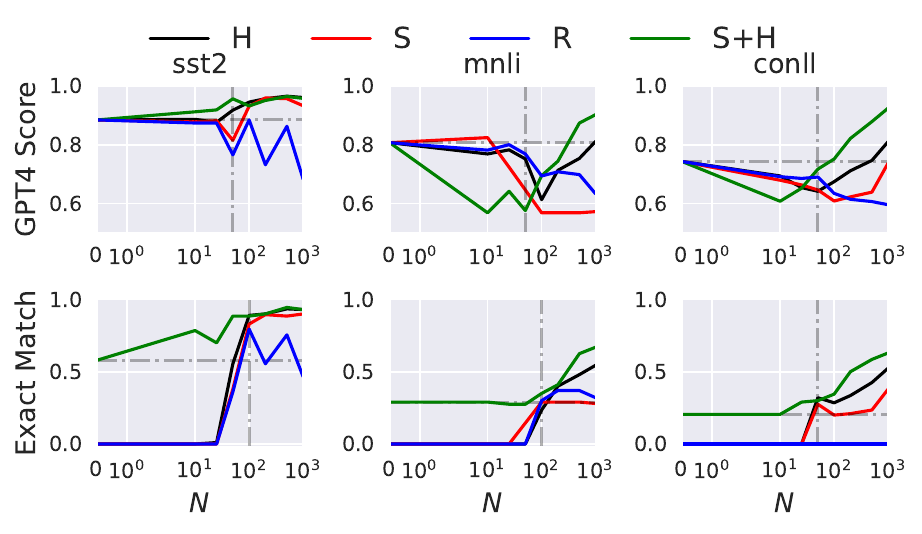}
    \caption{Incorporating  $0 \leq N \leq 1000$ (H)uman, (S)ynthetic and (R)andomly labeled synthetic data samples in IFT training set. (S+H) is trained on 100 synthetic samples, then $N$ human data samples. }\vspace{-.8cm}
    \label{fig:random}
\end{figure}

\noindent \textbf{Leveraging Synthetic Data to Learn to Format.} Our findings suggest a straightforward approach to optimizing the use of real examples: \emph{employ synthetic examples to assist the model in mastering the desired format before relying on real samples to enhance overall model performance}. We repeat the previous experiment, replacing the $N$  human-annotated target task training samples (\emph{H}), by GPT4 synthetically-generated samples  (\emph{S}), or synthetic samples with random labels (\emph{R}) (\autoref{fig:random}) \texttt{Exact Match} shows synthetic or randomly labeled data can indeed be used to learn the desired format, although the better quality human data eventually yields better results with more samples. In (\emph{S+H}), we train on 100 synthetic samples, then on $N$ human-annotated samples. This technique enables the model to master the format before being trained on high-quality data, largely improving human annotated data sample efficiency.


\subsection{$\mathcal{S}_2$: IFT models as Task-Specific Solvers}

\begin{figure}[ht]
    \centering
    \includegraphics[width=0.4\textwidth]{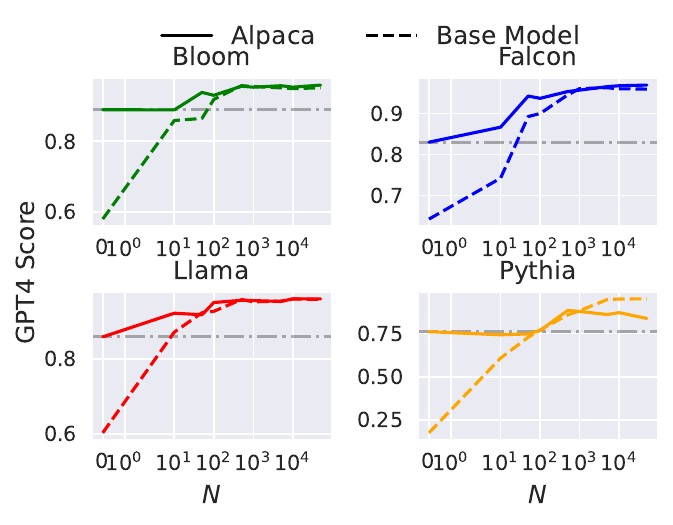}
    \caption{GPT4 score on SST-2 test set after finetuning with $0 \leq N \leq 1000$  samples on a (base) LM or an IFT model. Further experiments can be found in \autoref{sec:extra-datasets-exp4}. }
    \label{fig:exp4}
\end{figure}

\noindent \textbf{Setting.} We use four model architectures and, for each architecture, we employ the base model to train an IFT model variant using the synthetic Alpaca dataset. We then fine-tune both the base models and their IFT variants on a subset of $N$ samples drawn from the target task. This setup simulates an industrial scenario in which limited data is available to specialize a model on a unique task, and assesses whether there are benefits to instruction-tuning a base model before fine-tuning it on the target task. 
\\\noindent \textbf{Results}. \autoref{fig:exp4} demonstrates that IFT models exhibit enhanced performance in low-data scenarios (when $10 \leq N \leq 200$). 
Intuitively, IFT models are better able to leverage the task description given in the prompts, thus enabling boosted zero-shot performance \cite{scao2021data}. This complements and steers model training when finetuned with small numbers of samples. When more samples are available ($N \geq 200$), the task pattern is sufficiently clear and the benefits of prompting and of the IFT training phase disappear.

This finding aligns with the results presented in \autoref{ssec:scenario1}, emphasizing the potential of synthetic datasets to enhance data efficiency in industrial scenarios.

\section{Key Takeaways for Practitioners}
\label{sec:conclusion}

\noindent \textbf{Leveraging LLM for evaluation.} Evaluating IFT models is challenging, as it mandates \emph{comparability across tasks} and \emph{format-agnostism}, which standard metrics struggle with. While LLM scoring is not ideal (Limitations in \autoref{sec:limitations}), it is a strong option practitioners should add to their arsenal.
\\\noindent \textbf{Leveraging Synthetic Data for Efficient Learning.} LLM-based evaluation uncovers the fact that leveraging synthetic data provides a quick and cost-effective approach to mastering format in low data regimes, with no performance degradation. This methodology proves viable across various scenarios, presenting an opportunity to more efficiently leverage potentially limited amounts of expert annotated data available.






\section*{Limitations}
\label{sec:limitations}
While this paper has argued in favor of using LLM as scorers, important drawbacks remain. 
The best-performing scorer at the moment, GPT4 is a proprietary, black-box access model, and no guarantees exist that it will remain accessible unchanged over time, leading to reproducibility issues and data privacy concerns. Since model and training data internals are not open-knowledge, analysis of scoring errors and internal model biases is also limited. 

Promising openly available alternative models are being developed, either general purpose LLMs aiming to disrupt the hegemony of GPT4 \cite{touvron2023llama2, bai2023qwen}, or smaller models specialized for automatic evaluation, often attempting to distillate GPT4's scoring abilities by training on GPT4 generated scores or scoring explanations \cite{xu2023instructscore, liu2023geval}. In the latter category, the Prometheus scoring model \cite{kim2023prometheus}, based on Llama2, claims scoring performances on par with GPT4 in terms of human score correlations over a variety of tasks and benchmarks. Eventually, strong Open-Source LLMs should alleviate most of the concerns raised by relying on proprietary black-box models and we hope this work, by shedding light on the importance of LLM scoring, motivates these efforts to build open models with strong scoring abilities. 

\section*{Ethics Statement}

While this work intends to evaluate scorers across many different tasks and settings, it is essentially English-centric, and no conclusions are drawn about the robustness of LLM scorers in other languages. LLM scoring may also be affected by internal model bias acquired through pretraining or subsequent finetuning, and while efforts are made by OpenAI to mitigate bias, critical applications of LLM evaluation should consider that truly objective evaluation is not attainable.

All data and base models used in this work originate from publicly available sources. The GPT4 Alpaca dataset is a variant of \cite{alpaca2023} built from synthetic data only, collected through the OpenAI API. The non-synthetic data are sourced from manually annotated, widely used datasets for NLP benchmarking. This work does not transgress any usage restrictions associated with these data sources. Base models used are either available through fully open-source licenses (Falcon, Pythia), or licenses with no restrictions for research purposes (LLaMA, Bloom). 

We estimate our experiments consumed 5500 GPU V100 hours, using a low-carbon compute cluster, amounting to about 950 kg of CO2 over the course of the project.  To reduce the impact to a maximum, all runs are done through the efficient Low-Rank Adaptation training strategy \cite{hu2021lora}, and only trained adapter weights are stored to minimize bandwidth and memory usage. API calls to external APIs are cached to minimize redundancies.

\section*{Acknowledgements}
This work is partially supported by Illuin Technology, and by a grant from ANRT France.
This work was performed using HPC resources from GENCI–IDRIS (Grant 2023-AD011014185).
Secondary compute contributions were made on HPC resources (Grant 2023-AD103256) and (Grant 2023-AD101838).


\bibliography{references}
\bibliographystyle{acl_natbib}

\clearpage
\newpage
\appendix
\input{appendix.tex}

\end{document}

%% file: appendix.tex
\section{Ressources}

\subsection{Data}
The GPT4 Alpaca dataset is collected from \url{https://huggingface.co/datasets/vicgalle/alpaca-gpt4}. 

\label{sec:alpaca-pres}
We define a taxonomy of 13 subtasks that the instructions can fall into: Classify, Code, Answer from Context (Question Answering based on a given passage), Create (Artistically oriented Natural Langiage Generation), Extract, Logic (Reasoning tasks), Answer from Memory (Question Answering from internal model knowledge), Summarize from Memory (Summarization/Explanation from internal model knowledge), Rewrite (Reformulation tasks), Write (Natural Language Generation tasks with a non-artistic goal), Summarize (Summarization tasks given a passage), Translate, and Other.
We then tag each of the 52000 instructions with the subtask it falls into using GPT4 as an oracle by prompting it with the classification task.
To verify the quality of the tagging, we manually verify a random sample of 100 instructions and find a 93\% agreement rate. 

The other datasets also originate from the HuggingFace Hub \footnote{\url{https://huggingface.co/datasets/glue}}\footnote{\url{https://huggingface.co/datasets/squad_v2}}\footnote{\url{https://huggingface.co/datasets/conll2003}}. We format all tasks as instructions using the following prompts and concatenating the input at the end.

\paragraph{MNLI} \cite{N18-1101} Classify the following relationship between the Hypothesis sentence and the Premise sentence, as either Entailment, Contradiction or Neutral. 

\paragraph{QNLI} \cite{wang2019glue, rajpurkar2016squad} Classify whether the given context contains enough information to answer the question (answerable) or not (unanswerable).

\paragraph{STSB} \cite{agirre-etal-2013-sem} Give an integer score between 1 and 5, describing how similar sentence1 and sentence2 are. 5 means they are very similar, 1 means they are nothing alike.

\paragraph{SST2} \cite{socher2013recursive} Classify the following sentence as negative or positive.

\paragraph{CONLL} \cite{tjong-kim-sang-de-meulder-2003-introduction} Extract locations, persons, and organizations from the text. The output should be formatted as a JSON object with three keys: LOC (locations), PER (persons), and ORG (organizations). Each key should have a value that is a list of strings. If the text contains no entities of a given type, the corresponding list should be empty.

\paragraph{SQUADV2} \cite{rajpurkar2016squad} Answer the question depending on the context. You must only answer with one excerpt from the text.

\paragraph{XSUM} \cite{Narayan2018DontGM} Summarize the following article in a few words.

Finally we split each data task category into 3 sets: a training set, a validation set, and a test set, with 80\%, 10\%, and 10\% of the data respectively.

\subsection{Models}
\label{sec:models}

Models are publicly available on the HuggingFace Hub, and are fairly similar decoder GPT architectures. We select the 7 billion parameter version to compare models with similar scales: \textbf{LLaMA} \cite{touvron2023llama},  \textbf{Falcon} \cite{penedo2023refinedweb}, \textbf{Pythia} \cite{biderman2023pythia}, \textbf{Bloom} \cite{scao2022bloom}).

\subsection{Metrics}
\label{sec:metrics}

Model performance is measured using two family of methods: \textit{reference-based metrics} and \textit{LM metrics}.

\paragraph{Reference-Based metrics} are the common metrics used in the litterature to evaluate the performance of a generative model on a given task, and offer a measure of the distance between a generation and a reference output. In this category are included the \textbf{Exact Match} and F-measures, but also more task-specific metrics based on co-occurence of words betwwen the output and the reference, such as \textbf{ROUGE} scores \cite{lin-2004-rouge} for summarization, BLEU scores for translation \cite{papineni-etal-2002-bleu}. To go beyond word matching heuristics, we also baseline neural network scorers. \textbf{SentenceBert} models \cite{reimers2019sentencebert} enables to calculate the cosine similarity between sentence embeddings; in this work a general purpose embedding model (all-mini-lm-l6-v2) is used. \textbf{BertScore} computes a F1 score leveraging pairwise cosine similarity between tokens, using a strong encoder model finetuned on an NLI task as shown to be best, here we use \texttt{microsoft/deberta-base-mnli} from the HuggingFace Hub \cite{he2023debertav3}.

\paragraph{LM metrics} are metrics that are computed by directly scoring the output using a language model. In this work, we experiment with \textbf{GPT4} and \textbf{GPT3.5} as scorers, by providing the evaluated model output and tasking the scorer to grade the quality of the output fom 0 to 10, based on relevance, fluency, factuality, coherence. The scoring prompts are task agnostic and allow to compare the performance of a model on different tasks, or on open-ended tasks where there is no ground truth and traditional literature metrics cannot be used. They also have the advantage of being continuous, which allows to study transitive regime without the "emergence" effect \cite{schaeffer2023emergent}. Finally, we baseline a \textbf{Reward Model (RM)} \cite{köpf2023openassistant}, a model trained on outputting a score designed to reflect a human's appreciation of a judgement. While this can be used as a reference-free scorer, we find increased robustness when reporting the score softmaxed with the RM score of the reference.

\paragraph{GPT Scoring prompt}
To obtain GPT4 and GPT3.5 scores for model responses to a given instruction, we prompt the OpenAI API as such:

\begin{quote}
You are a helpful assistant that helps evaluate the quality of two responses to a prompt.

Answer by awarding a score between 0 and 10 to each response, where 0 means the response is completely inappropriate and 10 means the response is very good.
A response that is acceptable should never be awarded less than 6 out of 10.

Answer base on the following criteria:\\
1. Is the response grammatically correct?\\
2. Is the response semantically correct?\\
3. Is the response coherent?\\
4. Is the response relevant to the prompt?\\

Output format (csv):\\
<score1 from 0 to 10>,<score2 from 0 to 10>\\

Rate the responses to the following instruction.\\
\{prompt\}

Response 1: 
\{response1\}

Response 2: 
\{response2\}

Output:
\end{quote}

Response 1 corresponds to the model prediction, and reference 2 to the "gold" label. The scores obtained are then checked for conformity (correct output template, scores between 0 and 10). Finally, they are scaled between 0 and 1.

\subsection{Framework}
Code is written using PyTorch and the Transformers library, as well as the PEFT library to train models using  low-rank adaptation. Training runs are done on compute clusters with NVIDIA V100 32GB GPUs.

\subsection{Default Model Training}
\label{sec:default-setup}
 We train models across a wide range of models, training data source and dataset sizes. To stay consistent between runs, we train models for 400 steps with a 128 batch size, achieved through gradient accumulation, a 5e-4 learning rate with linear decay, a warmup of 100 steps, or one epoch (whichever is lowest). We frequently log validation split CE loss and use early stopping to prevent overfitting. Code is fully released at \url{https://github.com/ManuelFay/IFTEval}. Training runs to completion take about 5 hours on a V100 GPU and 2h on a A100 GPU. This default setup is used to train all models in this work.

The only exception are models trained on datasets with less than 128 samples (less than the batch size). In these cases, we select a batch size of 8.

\section{Re-evaluating evaluation}

\subsection{Experimental setup}
\label{sec:exp_setup_1}

\textbf{Summarization dataset} In this dataset \cite{stiennon2022learning}, summaries are already generated and associated with human scores. We select 4 summarization models from the dataset, and 200 inputs the 4 models have been evaluated in common on, and that have at least 2 different human annotators. This enables computing Spearman Rank Correlation both between metrics and the mean human annotation, but also between human annotators by averaging the pairwise human correlation scores over all pairs of annotators.

\textbf{Synthetic Alpaca dataset}  Models are trained in the default conditions explicited in \ref{sec:default-setup}. To construct Spearman rank correlations between evaluators, we train a suite of over 52 models in which the training data mixture slightly differs (exclusion or partial exclusion of a target category). To do so, we (i) select one of the 13 task categories identified in \ref{sec:alpaca-pres}. (ii) We then build a base training data mixtures by concatenating the training splits from all other categories, respectively the validation splits. (iii) Finally, we randomly sample respectively 0, 10, 100 and 1000 samples from the held-out category's training split and add them to the base mixture. (v) We train one model per data mixture according to the training guidelines in \ref{sec:default-setup}. (v) We evaluate performance with all metrics on the held-out task test split. (vi) The process is repeated for all categories.

\textbf{Human-annotated datasets} Models are trained in the default conditions explicited in \ref{sec:default-setup}. To evaluate Spearman correlation between rankers (CAT and CIT scenarios), we train a suite of over 300 models in which the training data mixture slightly differs (exclusion or partial exclusion of a target category). To do so, we (i) select all 13 task categories identified in \ref{sec:alpaca-pres}. (ii) We then build a base training data mixtures by concatenating the training splits from all categories, respectively the validation splits. (iii) We then select a category from the benchmark tasks (MNLI, QNLI, STSB, SST2, Squad, XSum, CONLL). (iv) We randomly sample respectively 0, 10, 25, 50, 100, 200, 500 and 1000 samples from the selected category's training split and add them to the base mixture, and repeat the sampling process 4 times;, to obtain 7*8*4 = 504 data mixtures. (iv) We train one model per data mixture according to the training guidelines in \ref{sec:default-setup}. (v) We evaluate performance with GPT4 on the target task test split. (vi) To compute the Spearman correlations, we aim to reduce dataset specific formatting artefacts and thus only select models trained with less than 100 target task samples in the training mixture. Correlations are computed two ways, either by computing Spearman correlations per task category and averaging the correlation values with human judgement (CIT), or by computing correlations globally without considering task category information (CAT) enabling the study of cross-task scoring comparability. 

\textbf{TFA}
Finally, to compute the \texttt{TFA}, that is the relative performance improvements after format is learned, we select 3 target tasks, MNLI, QNLI, SST2. These datasets are the three datasets we have perfect sample-wise binary human evaluation for from the set of \emph{human data} benchmarks. From the suite of trained models, we select 5 base models finetuned on the Alpaca dataset, and for each task, 5 models trained on both Alpaca and 1000 samples of the target task. Previous experiments (\autoref{fig:exp1b_all}) have shown that on these datasets, format is generally learned after a few hundred samples. This enables us to compare models that learned target task output format with models that have not, using the various metrics at our disposal. We compute the relative performance improvement of all metrics, and average them in Tab. \ref{tab:main_metrics} (\texttt{TSA}). While human evaluation shows model performance has progressed with the introduction of target task samples (12\%), non-LLM metrics largely overestimate the boost in response performance, because of their strong bias towards outputs matching the "gold" reference. In low-shot settings when no formatting is learned, it is particularly interesting to use format agnostic scorers like GPT4 to truly analyze a model's comprehension of a task.

\subsection{Obtaining Human Scores on the benchmark data} 

To facilitate the collection of human annotated model outputs on the non-synthetic data, we simplify the problem by reducing it to a binary choice (Correct / Incorrect)  on classification-like datasets (SST2, MNLI, QNLI). We manually observe outputs for each trained model and output task category, and craft pre-tagging heuristics to make the annotation process quicker. These heuristics are empirically built with knowledge of model outputs; for example on zero-shot sentiment classification tasks, Llama models will often answer \emph{"The sentence is Positive"}, instead of \emph{"positive"}, but trivial heuristic functions can assist in tagging these as correct answers. This is facilitated by the fact output structure for a given model/task combination is often very similar (especially at low decoding temperatures), and we can iterate on these heuristic functions until no error is detected at all. For non-classification tasks, we adopt a strict rating scheme as well. NER responses through the CONLL 2003 task are considered correct only when the dictionary contains all correct key/value pairs with respect to the ground truth, but disregarding formatting artefacts. For SquadV2, an extractive task, we report the F1 intersection between the predicted answer span, and the correct answer, similarly as what is done in \cite{rajpurkar2016squad}. Finally, for summarization, we report ROUGE scores for XSUM as a reference point, but do not include these results with the other tasks during correlation computations, and rather compute human correlations using the human annotation dataset from \cite{stiennon2022learning}, for which we report results on Table  \ref{tab:main_metrics} (SUM).

Code is released at \url{https://github.com/ManuelFay/IFTEval}.

\subsection{Results on Alpaca test set}

The full results per category are reported in Table \ref{tab:exp1a}.

\begin{table*}
    \centering
    \begin{tabular}{lrrrrrrr}
        \toprule
         & GPT4 & GPT3.5 & ROUGE-1 & BERTScore & SBERT & RM & Soft RM \\
        \midrule
        Write & 0.97 & 0.86 & 0.41 & 0.72 & 0.74 & 3.21 & 0.33 \\
        Answer from Context & 0.94 & 0.86 & 0.55 & 0.76 & 0.80 & 1.29 & 0.45 \\
        Answer from Memory & 0.93 & 0.87 & 0.47 & 0.73 & 0.84 & 2.78 & 0.28 \\
        Extract & 0.92 & 0.88 & 0.54 & 0.77 & 0.83 & 1.61 & 0.52 \\
        Summarize from Memory & 0.92 & 0.85 & 0.45 & 0.73 & 0.84 & 3.21 & 0.43 \\
        Summarize from Context & 0.89 & 0.86 & 0.54 & 0.79 & 0.84 & 1.47 & 0.42 \\
        Rewrite & 0.89 & 0.87 & 0.68 & 0.86 & 0.82 & -0.20 & 0.47 \\
        Translate & 0.88 & 0.94 & 0.58 & 0.81 & 0.71 & 1.78 & 0.44 \\
        Create & 0.85 & 0.78 & 0.34 & 0.67 & 0.62 & -0.20 & 0.27 \\
        Classify & 0.85 & 0.76 & 0.50 & 0.73 & 0.72 & 0.74 & 0.43 \\
        Code & 0.79 & 0.82 & 0.46 & 0.71 & 0.74 & 0.73 & 0.43 \\
        Logic Reasoning & 0.71 & 0.88 & 0.52 & 0.73 & 0.80 & 0.33 & 0.49 \\
        \bottomrule
        \end{tabular}
    \caption{Averaged score over samples and models of each metric on each category. Category \textit{Other} is not listed as it  consists in synthetic instructions often not answerable by LLMs without external tooling.}
    \label{tab:exp1a}
\end{table*}

\subsection{Automatic Error Analysis}

To go further in the analysis and uncover novel insights at large scale, LLM models can for example enable automatic error analysis. We select a few low scoring outputs and prompt the scoring models to detect mistake patterns, or give potential explanations for the generative model's shortcomings On the sentiment classification task, automatic error analysis correctly detects that most errors are false positives, triggered by sarcasm, or "the presence of words that might be associated with positive sentiment in other contexts". For NLI tasks, it detects the tested model has troubles with  with "understanding negations, contradictions, and subtle differences in meaning between the premise and hypothesis".

\paragraph{Sentiment Analysis}
The following is a series of predictions a sentiment classification ML model got wrong. Can you spot any patterns in the types of mistakes the model made and explain in what areas the model needs to improve?

\begin{quote}

Sentence:
the château would have been benefited from a sharper , cleaner script before it went in front of the camera . 
Model prediction: The sentence is a positive statement.
Ground truth: negative

Sentence:
but there 's plenty to offend everyone ... 
Model prediction: The sentence is positive.
Ground truth: negative

Sentence:
outtakes in which most of the characters forget their lines and just utter ` uhhh , ' which is better than most of the writing in the movie 
Model prediction: The sentence is positive.
Ground truth: negative

Sentence:
the filmmakers ' paws , sad to say , 
Model prediction: The sentence is positive.
Ground truth: negative

Sentence:
in the media 
Model prediction: The sentence "in the media" is a positive statement.
Ground truth: negative

Sentence:
profanities 
Model prediction: The sentence "profanities" is a positive statement.
Ground truth: negative

\end{quote}

'The model seems to struggle with understanding the context and sarcasm in the sentences. It often misinterprets negative statements as positive, possibly due to the presence of words that might be associated with positive sentiment in other contexts. The model needs to improve in understanding the overall context of the sentences and detecting sarcasm or irony. Additionally, it seems to struggle with shorter sentences or single-word inputs, where it might not have enough information to accurately determine the sentiment.'

\paragraph{NLI}
The following is a series of predictions a NLI ML model got wrong. Can you spot any patterns in the types of mistakes the model made and explain in what areas the model needs to improve?

\begin{quote}
Premise:
He pulled his cloak tighter and wished for a moment that he had not shaved his head.
Hypothesis:
The man pulled his super hero cape around himself to show off.
Model prediction: Entailment.
Ground truth: neutral

Premise:
because i always had to do it and so i just pay someone else to do it and they do the they do the cutting they fertilize they um edge and um i think this year i'm going to have some landscaping put in
Hypothesis:
I still do all the gardening and landscaping myself. 
Model prediction: Entailment.
Ground truth: contradiction

Premise:
He's chosen Meg Ryan.
Hypothesis:
Jon Doe was chosen.
Model prediction: Entailment.
Ground truth: contradiction

Premise:
yep that's what he's worried about the trees or a bush because lilac bushes they they grow fast some people uh would really like to have them and then the people that do have them they spread and they sprout all over their their lawn
Hypothesis:
He's not worried about the trees. Lilac bushes take a long time to grow.
Model prediction: Entailment.
Ground truth: contradiction

Premise:
Trying Your Luck
Hypothesis:
This is not a game of luck, but one of pure skill.
Model prediction: Entailment.
Ground truth: contradiction

Premise:
He bent down to study the tiny little jeweled gears.
Hypothesis:
The were no jewels on the gears he examined.
Model prediction: Entailment.
Ground truth: contradiction

Premise:
As a basic guide, the symbols below have been used to indicate high-season rates in Hong Kong dollars, based on double occupancy, with bath or shower.
Hypothesis:
This page intentionally left blank.
Model prediction: Entailment.
Ground truth: contradiction

Premise:
In fact, European nations need to do some serious fiscal housecleaning.
Hypothesis:
There needs to be some serious fiscal housecleaning by European nations.
Model prediction: The relationship between the Premise sentence and the Hypothesis sentence is Neutral. The two sentences are not contradictory or entailing.
Ground truth: entailment

Premise:
It means that they gather and interpret their material fairly and argue about its interpretations rationally.
Hypothesis:
They gather and interpret the material fairly and argue rationally about it.
Model prediction: The relationship between the Premise sentence and the Hypothesis sentence is Neutral. Both sentences are independent of each other and do not entail or contradict each other.
Ground truth: entailment

Premise:
well that's pretty typical though uh i don't uh i don't guess it's going to be any much different uh than than it has been in the past so i expect uh July and August we'll see our  or uh share of hundred degree days
Hypothesis:
We don't ever see the temperature get into the hundreds.
Model prediction: Entailment.
Ground truth: contradiction

Premise:
They would burn to the ground by morning.
Hypothesis:
They would not burn to the ground by morning.
Model prediction: Entailment.
Ground truth: contradiction

Premise:
A rusty iron gate swinging dismally on its hinges! 
Hypothesis:
The iron gate did not move at all.  
Model prediction: Entailment.
Ground truth: contradiction

\end{quote}

The model seems to have a pattern of incorrectly predicting entailment when the ground truth is either neutral or contradiction. It appears to struggle with understanding negations, contradictions, and subtle differences in meaning between the premise and hypothesis. The model also seems to have difficulty with understanding the context and relationships between entities mentioned in the sentences.
To improve, the model needs to focus on better understanding negations, identifying contradictions, and recognizing subtle differences in meaning and context between sentences. Additionally, it should work on better identifying the relationships between entities and the overall context of the sentences.

\subsection{Correlations on the non-synthetic human tasks}

Table \ref{tab:exp1b_corr} and Table \ref{tab:exp1b_mean} display the scorer correlations between themselves on the non-synthetic datasets in both the \texttt{CAT} setting (Table \ref{tab:exp1b_corr}), and the \texttt{CIT} setting (Table \ref{tab:exp1b_mean}). Results are consistent with \autoref{fig:correlations}, with BertScore, SentenceBert, Rouge all displaying greater correlations between themselves than with reference-free metrics, and GPT4 standing out both on scoring within each tasks (CIT) and between tasks (CAT).

\begin{table*}
    \centering
    \begin{tabular}{lrrrrrrr}
    \toprule
     & ROUGE-1 & GPT4 & GPT3.5 & BScore & SBert & RM & Human \\
    \midrule
    ROUGE-1 & 1.00 & -0.16 & 0.46 & 0.73 & 0.85 & -0.77 & 0.22 \\
    GPT4 & -0.16 & 1.00 & -0.28 & 0.10 & -0.04 & 0.56 & 0.68 \\
    GPT3.5 & 0.46 & -0.28 & 1.00 & -0.00 & 0.16 & -0.37 & -0.19 \\
    BScore & 0.73 & 0.10 & -0.00 & 1.00 & 0.91 & -0.61 & 0.22 \\
    SBert & 0.85 & -0.04 & 0.16 & 0.91 & 1.00 & -0.69 & 0.29 \\
    RM & -0.77 & 0.56 & -0.37 & -0.61 & -0.69 & 1.00 & 0.28 \\
    Human & 0.22 & 0.68 & -0.19 & 0.22 & 0.29 & 0.28 & 1.00 \\
    \bottomrule
    \end{tabular}
    \caption{\textbf{Spearman $\rho$ between scorers} on non-synthetic data in the CAT setting.}
    \label{tab:exp1b_corr}
\end{table*}

\begin{table*}
    \centering
    \begin{tabular}{lrrrrrrr}
    \toprule
     & ROUGE-1 & GPT4 & GPT3.5 & BScore & SBert & RM & Human \\
    \midrule
    ROUGE-1 & 1.00 & 0.50 & 0.33 & 0.63 & 0.90 & -0.10 & 0.57 \\
    GPT4 & 0.50 & 1.00 & 0.66 & 0.07 & 0.35 & 0.40 & 0.77 \\
    GPT3.5 & 0.33 & 0.66 & 1.00 & -0.10 & 0.16 & 0.24 & 0.48 \\
    BScore & 0.63 & 0.07 & -0.10 & 1.00 & 0.66 & -0.43 & 0.13 \\
    SBert & 0.90 & 0.35 & 0.16 & 0.66 & 1.00 & -0.18 & 0.43 \\
    RM & -0.10 & 0.40 & 0.24 & -0.43 & -0.18 & 1.00 & 0.29 \\
    Human & 0.57 & 0.77 & 0.48 & 0.13 & 0.43 & 0.29 & 1.00 \\
    \bottomrule
    \end{tabular}
    \caption{\textbf{Spearman $\rho$ between metrics} on non-synthetic data in the CIT setting.}
    \label{tab:exp1b_mean}
\end{table*}

\section{IFT Models For Industrial Applications}

\subsection{$\mathcal{S}_1$: Improving Specific Tasks}

\subsubsection{Experimental setup}
\label{sec:exp_setup_app_31}
Models are trained in the default conditions explicited in \autoref{sec:default-setup}. To evaluate target task performance as target task samples are progressively introduced in the generalist synthetic training set, we train a suite, we (i) select all 13 task categories identified in \autoref{sec:alpaca-pres}. (ii) We then build a base training data mixtures by concatenating the training splits from all categories, respectively the validation splits. (iii) We then select a category from the benchmark tasks (MNLI, QNLI, STSB, SST2, SquadV2, Xsum, CONLL). (iv) We randomly sample respectively 0, 10, 25, 50, 100, 200, 500 and 1000 samples from the selected category's training split and add them to the base mixture, and repeat the sampling process 4 times;, to obtain 7*8*4 = 504 data mixtures. (iv) We train one model per data mixture according to the training guidelines in \autoref{sec:default-setup}. (v) We evaluate performance with all metrics listed in \autoref{sec:metrics} on the target task test split.

\begin{figure}[ht]
    \centering
    \includegraphics[width=0.5\textwidth]{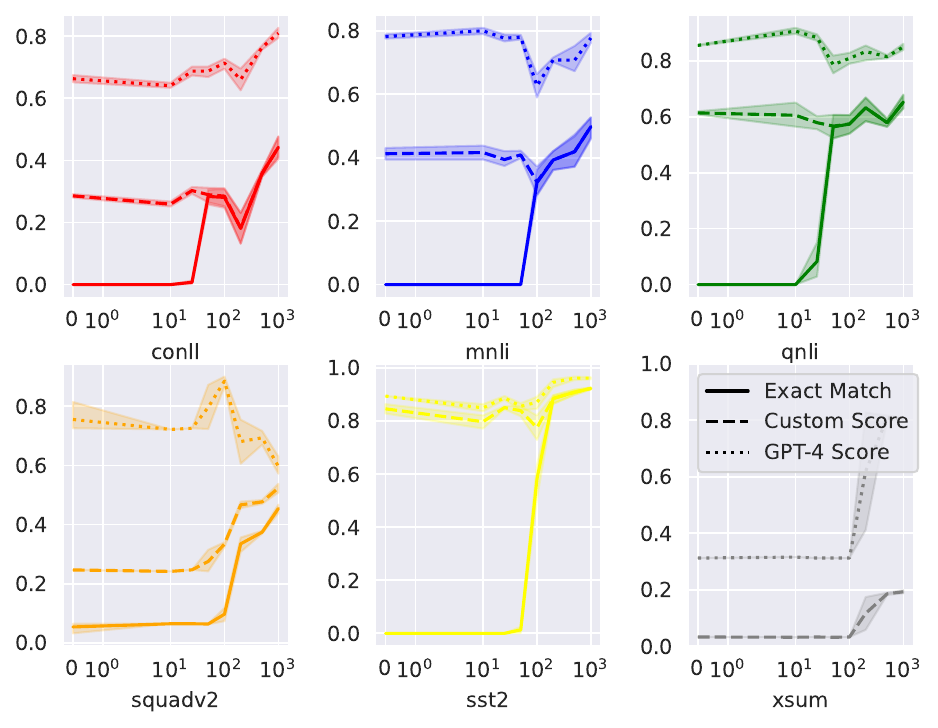}
    \caption{LLaMA performance after incorporating $0 \leq N \leq 1000$ real task samples from respectively CONLL, MNLI, QNLI, SquadV2, SST2 and XSUM into the base training mixture}
    \label{fig:exp2}
\end{figure}

\subsubsection{Performance degradation}
\label{sec:app_perf_deg}
Full performance degradation results after 1000 samples of task-specific data are integrated within the synthetic test set are shown in \autoref{tab:perf_deg}.

\begin{table}
    \centering
    \begin{tabular}{lrrrrr}
    \toprule
     & Base & sst2 & conll & mnli & Average \\
    \midrule
    Bloom & 0.75 & 0.74 & 0.74 & 0.73 & 0.73 \\
    Falcon & 0.87 & 0.86 & 0.87 & 0.84 & 0.85 \\
    Llama & 0.84 & 0.85 & 0.86 & 0.86 & 0.86 \\
    Pythia & 0.67 & 0.66 & 0.66 & 0.66 & 0.66 \\
    \bottomrule
    \end{tabular}
    
    \caption{Performance degradation on the Alpaca test set, after introducing 1000 samples of specialized data to the train set. Base denotes the reference scores, computed from the initial zero-shot performance.}
    \label{tab:perf_deg}
\end{table}

\subsection{$\mathcal{S}_2$: IFT models as Task-Specific Solvers}

\subsubsection{Experimental setup}
Models are trained in the default conditions explicited in \ref{sec:default-setup}. Starting from either a base model, or an instruction fine-tuned model, we evaluate target task performance as the model is fine-tuned solely on a varying number of target task training samples. We (i) select a target task from the real data benchmark tasks. (ii) We then sample a set of $N \in \{0, 10, 25, 50, 100, 200, 500, 1000\}$ samples from the training set, and pick 100 samples to serve as a validation split. (iii) We repeat the sampling process 4 times to mitigate variations between runs. (iv) We train one model per data mixture according to the training guidelines in \ref{sec:default-setup}. (v) We evaluate performance with all metrics listed in \ref{sec:metrics} on the target task test split.

\subsubsection{Extra models}
\label{sec:extra-datasets-exp4}
\begin{figure}[ht]
    \centering
    \includegraphics[width=0.5\textwidth]{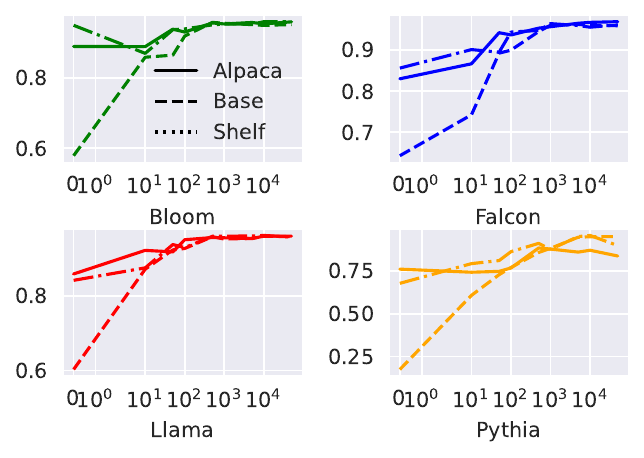}
    \caption{GPT4 score on SST-2 test set after finetuning with $0 \leq N \leq 1000$  samples on a (base) LM, an "off-the-shelf" IFT model, and an IFT model fine-tuned on Alpaca.}
    \label{fig:exp4_old}
\end{figure}

To evaluate the impact of IFT quality of the findings of the $\mathcal{S}_2$ experiment, we repeat the experiment using an "off-the-shelf" strong IFT model from the HuggingFace Hub. Notably, we select the 7B variant of \cite{vicuna2023} for the Llama variant, BloomZ for Bloom \cite{muennighoff2022crosslingual}, Falcon-Instruct for Falcon \cite{penedo2023refinedweb} and an IFT version of Pythia trained by \cite{köpf2023openassistant}. Results reported in Figure \ref{fig:exp4_old} show similar dynamics between the off-the-shelf models and the variants Instruction Fine-tuned in this work for one epoch on the Alpaca dataset \cite{alpaca2023}.

%% file: main.bbl
\begin{thebibliography}{57}
\expandafter\ifx\csname natexlab\endcsname\relax\def\natexlab#1{#1}\fi

\bibitem[{Agirre et~al.(2013)Agirre, Cer, Diab, Gonzalez-Agirre, and
  Guo}]{agirre-etal-2013-sem}
Eneko Agirre, Daniel Cer, Mona Diab, Aitor Gonzalez-Agirre, and Weiwei Guo.
  2013.
\newblock \href {https://aclanthology.org/S13-1004} {*{SEM} 2013 shared task:
  Semantic textual similarity}.
\newblock In \emph{Second Joint Conference on Lexical and Computational
  Semantics (*{SEM}), Volume 1: Proceedings of the Main Conference and the
  Shared Task: Semantic Textual Similarity}, pages 32--43, Atlanta, Georgia,
  USA. Association for Computational Linguistics.

\bibitem[{Bai et~al.(2023)Bai, Bai, Chu, Cui, Dang, Deng, Fan, Ge, Han, Huang,
  Hui, Ji, Li, Lin, Lin, Liu, Liu, Lu, Lu, Ma, Men, Ren, Ren, Tan, Tan, Tu,
  Wang, Wang, Wang, Wu, Xu, Xu, Yang, Yang, Yang, Yang, Yao, Yu, Yuan, Yuan,
  Zhang, Zhang, Zhang, Zhang, Zhou, Zhou, Zhou, and Zhu}]{bai2023qwen}
Jinze Bai, Shuai Bai, Yunfei Chu, Zeyu Cui, Kai Dang, Xiaodong Deng, Yang Fan,
  Wenbin Ge, Yu~Han, Fei Huang, Binyuan Hui, Luo Ji, Mei Li, Junyang Lin, Runji
  Lin, Dayiheng Liu, Gao Liu, Chengqiang Lu, Keming Lu, Jianxin Ma, Rui Men,
  Xingzhang Ren, Xuancheng Ren, Chuanqi Tan, Sinan Tan, Jianhong Tu, Peng Wang,
  Shijie Wang, Wei Wang, Shengguang Wu, Benfeng Xu, Jin Xu, An~Yang, Hao Yang,
  Jian Yang, Shusheng Yang, Yang Yao, Bowen Yu, Hongyi Yuan, Zheng Yuan,
  Jianwei Zhang, Xingxuan Zhang, Yichang Zhang, Zhenru Zhang, Chang Zhou,
  Jingren Zhou, Xiaohuan Zhou, and Tianhang Zhu. 2023.
\newblock \href {http://arxiv.org/abs/2309.16609} {Qwen technical report}.

\bibitem[{Bhandari et~al.(2020)Bhandari, Gour, Ashfaq, Liu, and
  Neubig}]{bhandari-etal-2020-evaluating}
Manik Bhandari, Pranav~Narayan Gour, Atabak Ashfaq, Pengfei Liu, and Graham
  Neubig. 2020.
\newblock \href {https://doi.org/10.18653/v1/2020.emnlp-main.751}
  {Re-evaluating evaluation in text summarization}.
\newblock In \emph{Proceedings of the 2020 Conference on Empirical Methods in
  Natural Language Processing (EMNLP)}, pages 9347--9359, Online. Association
  for Computational Linguistics.

\bibitem[{Biderman et~al.(2023)Biderman, Schoelkopf, Anthony, Bradley, O'Brien,
  Hallahan, Khan, Purohit, Prashanth, Raff, Skowron, Sutawika, and van~der
  Wal}]{biderman2023pythia}
Stella Biderman, Hailey Schoelkopf, Quentin Anthony, Herbie Bradley, Kyle
  O'Brien, Eric Hallahan, Mohammad~Aflah Khan, Shivanshu Purohit, USVSN~Sai
  Prashanth, Edward Raff, Aviya Skowron, Lintang Sutawika, and Oskar van~der
  Wal. 2023.
\newblock \href {http://arxiv.org/abs/2304.01373} {Pythia: A suite for
  analyzing large language models across training and scaling}.

\bibitem[{Bubeck et~al.(2023)Bubeck, Chandrasekaran, Eldan, Gehrke, Horvitz,
  Kamar, Lee, Lee, Li, Lundberg, Nori, Palangi, Ribeiro, and
  Zhang}]{bubeck2023sparks}
Sébastien Bubeck, Varun Chandrasekaran, Ronen Eldan, Johannes Gehrke, Eric
  Horvitz, Ece Kamar, Peter Lee, Yin~Tat Lee, Yuanzhi Li, Scott Lundberg,
  Harsha Nori, Hamid Palangi, Marco~Tulio Ribeiro, and Yi~Zhang. 2023.
\newblock \href {http://arxiv.org/abs/2303.12712} {Sparks of artificial general
  intelligence: Early experiments with gpt-4}.

\bibitem[{Chen et~al.(2022)Chen, Belouadi, and Eger}]{chen2022reproducibility}
Yanran Chen, Jonas Belouadi, and Steffen Eger. 2022.
\newblock Reproducibility issues for bert-based evaluation metrics.
\newblock \emph{arXiv preprint arXiv:2204.00004}.

\bibitem[{Chhun et~al.(2022)Chhun, Colombo, Clavel, and
  Suchanek}]{chhun2022human}
Cyril Chhun, Pierre Colombo, Chlo{\'e} Clavel, and Fabian~M Suchanek. 2022.
\newblock Of human criteria and automatic metrics: A benchmark of the
  evaluation of story generation.
\newblock \emph{arXiv preprint arXiv:2208.11646}.

\bibitem[{Chiang et~al.(2023)Chiang, Li, Lin, Sheng, Wu, Zhang, Zheng, Zhuang,
  Zhuang, Gonzalez, Stoica, and Xing}]{vicuna2023}
Wei-Lin Chiang, Zhuohan Li, Zi~Lin, Ying Sheng, Zhanghao Wu, Hao Zhang, Lianmin
  Zheng, Siyuan Zhuang, Yonghao Zhuang, Joseph~E. Gonzalez, Ion Stoica, and
  Eric~P. Xing. 2023.
\newblock \href {https://lmsys.org/blog/2023-03-30-vicuna/} {Vicuna: An
  open-source chatbot impressing gpt-4 with 90\%* chatgpt quality}.

\bibitem[{Colombo et~al.(2022{\natexlab{a}})Colombo, Noiry, Irurozki, and
  Cl{\'e}men{\c{c}}on}]{colombo2022best}
Pierre Colombo, Nathan Noiry, Ekhine Irurozki, and St{\'e}phan
  Cl{\'e}men{\c{c}}on. 2022{\natexlab{a}}.
\newblock What are the best systems? new perspectives on nlp benchmarking.
\newblock \emph{Advances in Neural Information Processing Systems},
  35:26915--26932.

\bibitem[{Colombo et~al.(2021)Colombo, Staerman, Clavel, and
  Piantanida}]{colombo2021automatic}
Pierre Colombo, Guillaume Staerman, Chlo{\'e} Clavel, and Pablo Piantanida.
  2021.
\newblock Automatic text evaluation through the lens of wasserstein
  barycenters.
\newblock \emph{arXiv preprint arXiv:2108.12463}.

\bibitem[{Colombo et~al.(2022{\natexlab{b}})Colombo, Clavel, and
  Piantanida}]{colombo2022infolm}
Pierre Jean~A Colombo, Chlo{\'e} Clavel, and Pablo Piantanida.
  2022{\natexlab{b}}.
\newblock Infolm: A new metric to evaluate summarization \& data2text
  generation.
\newblock In \emph{Proceedings of the AAAI Conference on Artificial
  Intelligence}, volume~36, pages 10554--10562.

\bibitem[{Dubois et~al.(2023)Dubois, Li, Taori, Zhang, Gulrajani, Ba, Guestrin,
  Liang, and Hashimoto}]{dubois2023alpacafarm}
Yann Dubois, Xuechen Li, Rohan Taori, Tianyi Zhang, Ishaan Gulrajani, Jimmy Ba,
  Carlos Guestrin, Percy Liang, and Tatsunori~B. Hashimoto. 2023.
\newblock \href {http://arxiv.org/abs/2305.14387} {Alpacafarm: A simulation
  framework for methods that learn from human feedback}.

\bibitem[{Fabbri et~al.(2021)Fabbri, Kry{\'s}ci{\'n}ski, McCann, Xiong, Socher,
  and Radev}]{fabbri-etal-2021-summeval}
Alexander~R. Fabbri, Wojciech Kry{\'s}ci{\'n}ski, Bryan McCann, Caiming Xiong,
  Richard Socher, and Dragomir Radev. 2021.
\newblock \href {https://doi.org/10.1162/tacl_a_00373} {{S}umm{E}val:
  Re-evaluating summarization evaluation}.
\newblock \emph{Transactions of the Association for Computational Linguistics},
  9:391--409.

\bibitem[{Gao et~al.(2021)Gao, Tow, Biderman, Black, DiPofi, Foster, Golding,
  Hsu, McDonell, Muennighoff, Phang, Reynolds, Tang, Thite, Wang, Wang, and
  Zou}]{eval-harness}
Leo Gao, Jonathan Tow, Stella Biderman, Sid Black, Anthony DiPofi, Charles
  Foster, Laurence Golding, Jeffrey Hsu, Kyle McDonell, Niklas Muennighoff,
  Jason Phang, Laria Reynolds, Eric Tang, Anish Thite, Ben Wang, Kevin Wang,
  and Andy Zou. 2021.
\newblock \href {https://doi.org/10.5281/zenodo.5371628} {A framework for
  few-shot language model evaluation}.

\bibitem[{Gudibande et~al.(2023)Gudibande, Wallace, Snell, Geng, Liu, Abbeel,
  Levine, and Song}]{gudibande2023false}
Arnav Gudibande, Eric Wallace, Charlie Snell, Xinyang Geng, Hao Liu, Pieter
  Abbeel, Sergey Levine, and Dawn Song. 2023.
\newblock \href {http://arxiv.org/abs/2305.15717} {The false promise of
  imitating proprietary llms}.

\bibitem[{He et~al.(2023)He, Gao, and Chen}]{he2023debertav3}
Pengcheng He, Jianfeng Gao, and Weizhu Chen. 2023.
\newblock \href {http://arxiv.org/abs/2111.09543} {Debertav3: Improving deberta
  using electra-style pre-training with gradient-disentangled embedding
  sharing}.

\bibitem[{Hendrycks et~al.(2021)Hendrycks, Burns, Basart, Zou, Mazeika, Song,
  and Steinhardt}]{hendrycks2021measuring}
Dan Hendrycks, Collin Burns, Steven Basart, Andy Zou, Mantas Mazeika, Dawn
  Song, and Jacob Steinhardt. 2021.
\newblock \href {http://arxiv.org/abs/2009.03300} {Measuring massive multitask
  language understanding}.

\bibitem[{Himmi et~al.(2023)Himmi, Irurozki, Noiry, Clemencon, and
  Colombo}]{himmi2023towards}
Anas Himmi, Ekhine Irurozki, Nathan Noiry, Stephan Clemencon, and Pierre
  Colombo. 2023.
\newblock Towards more robust nlp system evaluation: Handling missing scores in
  benchmarks.
\newblock \emph{arXiv preprint arXiv:2305.10284}.

\bibitem[{Hu et~al.(2021)Hu, Shen, Wallis, Allen-Zhu, Li, Wang, Wang, and
  Chen}]{hu2021lora}
Edward~J. Hu, Yelong Shen, Phillip Wallis, Zeyuan Allen-Zhu, Yuanzhi Li, Shean
  Wang, Lu~Wang, and Weizhu Chen. 2021.
\newblock \href {http://arxiv.org/abs/2106.09685} {Lora: Low-rank adaptation of
  large language models}.

\bibitem[{Kim et~al.(2023)Kim, Shin, Cho, Jang, Longpre, Lee, Yun, Shin, Kim,
  Thorne, and Seo}]{kim2023prometheus}
Seungone Kim, Jamin Shin, Yejin Cho, Joel Jang, Shayne Longpre, Hwaran Lee,
  Sangdoo Yun, Seongjin Shin, Sungdong Kim, James Thorne, and Minjoon Seo.
  2023.
\newblock \href {http://arxiv.org/abs/2310.08491} {Prometheus: Inducing
  fine-grained evaluation capability in language models}.

\bibitem[{Kocmi and Federmann(2023)}]{kocmi2023large}
Tom Kocmi and Christian Federmann. 2023.
\newblock \href {http://arxiv.org/abs/2302.14520} {Large language models are
  state-of-the-art evaluators of translation quality}.

\bibitem[{Köpf et~al.(2023)Köpf, Kilcher, von Rütte, Anagnostidis, Tam,
  Stevens, Barhoum, Duc, Stanley, Nagyfi, ES, Suri, Glushkov, Dantuluri,
  Maguire, Schuhmann, Nguyen, and Mattick}]{köpf2023openassistant}
Andreas Köpf, Yannic Kilcher, Dimitri von Rütte, Sotiris Anagnostidis,
  Zhi-Rui Tam, Keith Stevens, Abdullah Barhoum, Nguyen~Minh Duc, Oliver
  Stanley, Richárd Nagyfi, Shahul ES, Sameer Suri, David Glushkov, Arnav
  Dantuluri, Andrew Maguire, Christoph Schuhmann, Huu Nguyen, and Alexander
  Mattick. 2023.
\newblock \href {http://arxiv.org/abs/2304.07327} {Openassistant conversations
  -- democratizing large language model alignment}.

\bibitem[{Li et~al.(2023)Li, Hammoud, Itani, Khizbullin, and
  Ghanem}]{li2023camel}
Guohao Li, Hasan Abed Al~Kader Hammoud, Hani Itani, Dmitrii Khizbullin, and
  Bernard Ghanem. 2023.
\newblock \href {http://arxiv.org/abs/2303.17760} {Camel: Communicative agents
  for "mind" exploration of large scale language model society}.

\bibitem[{Liang et~al.(2022)Liang, Bommasani, Lee, Tsipras, Soylu, Yasunaga,
  Zhang, Narayanan, Wu, Kumar, Newman, Yuan, Yan, Zhang, Cosgrove, Manning,
  Ré, Acosta-Navas, Hudson, Zelikman, Durmus, Ladhak, Rong, Ren, Yao, Wang,
  Santhanam, Orr, Zheng, Yuksekgonul, Suzgun, Kim, Guha, Chatterji, Khattab,
  Henderson, Huang, Chi, Xie, Santurkar, Ganguli, Hashimoto, Icard, Zhang,
  Chaudhary, Wang, Li, Mai, Zhang, and Koreeda}]{liang2022holistic}
Percy Liang, Rishi Bommasani, Tony Lee, Dimitris Tsipras, Dilara Soylu,
  Michihiro Yasunaga, Yian Zhang, Deepak Narayanan, Yuhuai Wu, Ananya Kumar,
  Benjamin Newman, Binhang Yuan, Bobby Yan, Ce~Zhang, Christian Cosgrove,
  Christopher~D. Manning, Christopher Ré, Diana Acosta-Navas, Drew~A. Hudson,
  Eric Zelikman, Esin Durmus, Faisal Ladhak, Frieda Rong, Hongyu Ren, Huaxiu
  Yao, Jue Wang, Keshav Santhanam, Laurel Orr, Lucia Zheng, Mert Yuksekgonul,
  Mirac Suzgun, Nathan Kim, Neel Guha, Niladri Chatterji, Omar Khattab, Peter
  Henderson, Qian Huang, Ryan Chi, Sang~Michael Xie, Shibani Santurkar, Surya
  Ganguli, Tatsunori Hashimoto, Thomas Icard, Tianyi Zhang, Vishrav Chaudhary,
  William Wang, Xuechen Li, Yifan Mai, Yuhui Zhang, and Yuta Koreeda. 2022.
\newblock \href {http://arxiv.org/abs/2211.09110} {Holistic evaluation of
  language models}.

\bibitem[{Lin(2004)}]{lin-2004-rouge}
Chin-Yew Lin. 2004.
\newblock \href {https://aclanthology.org/W04-1013} {{ROUGE}: A package for
  automatic evaluation of summaries}.
\newblock In \emph{Text Summarization Branches Out}, pages 74--81, Barcelona,
  Spain. Association for Computational Linguistics.

\bibitem[{Liu et~al.(2023)Liu, Iter, Xu, Wang, Xu, and Zhu}]{liu2023geval}
Yang Liu, Dan Iter, Yichong Xu, Shuohang Wang, Ruochen Xu, and Chenguang Zhu.
  2023.
\newblock \href {http://arxiv.org/abs/2303.16634} {G-eval: Nlg evaluation using
  gpt-4 with better human alignment}.

\bibitem[{Mishra et~al.(2022)Mishra, Khashabi, Baral, and
  Hajishirzi}]{mishra2022crosstask}
Swaroop Mishra, Daniel Khashabi, Chitta Baral, and Hannaneh Hajishirzi. 2022.
\newblock \href {http://arxiv.org/abs/2104.08773} {Cross-task generalization
  via natural language crowdsourcing instructions}.

\bibitem[{Muennighoff et~al.(2022)Muennighoff, Wang, Sutawika, Roberts,
  Biderman, Scao, Bari, Shen, Yong, Schoelkopf
  et~al.}]{muennighoff2022crosslingual}
Niklas Muennighoff, Thomas Wang, Lintang Sutawika, Adam Roberts, Stella
  Biderman, Teven~Le Scao, M~Saiful Bari, Sheng Shen, Zheng-Xin Yong, Hailey
  Schoelkopf, et~al. 2022.
\newblock Crosslingual generalization through multitask finetuning.
\newblock \emph{arXiv preprint arXiv:2211.01786}.

\bibitem[{Narayan et~al.(2018)Narayan, Cohen, and Lapata}]{Narayan2018DontGM}
Shashi Narayan, Shay~B. Cohen, and Mirella Lapata. 2018.
\newblock Don't give me the details, just the summary! topic-aware
  convolutional neural networks for extreme summarization.
\newblock \emph{ArXiv}, abs/1808.08745.

\bibitem[{Ouyang et~al.(2022)Ouyang, Wu, Jiang, Almeida, Wainwright, Mishkin,
  Zhang, Agarwal, Slama, Ray, Schulman, Hilton, Kelton, Miller, Simens, Askell,
  Welinder, Christiano, Leike, and Lowe}]{ouyang2022training}
Long Ouyang, Jeff Wu, Xu~Jiang, Diogo Almeida, Carroll~L. Wainwright, Pamela
  Mishkin, Chong Zhang, Sandhini Agarwal, Katarina Slama, Alex Ray, John
  Schulman, Jacob Hilton, Fraser Kelton, Luke Miller, Maddie Simens, Amanda
  Askell, Peter Welinder, Paul Christiano, Jan Leike, and Ryan Lowe. 2022.
\newblock \href {http://arxiv.org/abs/2203.02155} {Training language models to
  follow instructions with human feedback}.

\bibitem[{Papineni et~al.(2002)Papineni, Roukos, Ward, and
  Zhu}]{papineni-etal-2002-bleu}
Kishore Papineni, Salim Roukos, Todd Ward, and Wei-Jing Zhu. 2002.
\newblock \href {https://doi.org/10.3115/1073083.1073135} {{B}leu: a method for
  automatic evaluation of machine translation}.
\newblock In \emph{Proceedings of the 40th Annual Meeting of the Association
  for Computational Linguistics}, pages 311--318, Philadelphia, Pennsylvania,
  USA. Association for Computational Linguistics.

\bibitem[{Penedo et~al.(2023)Penedo, Malartic, Hesslow, Cojocaru, Cappelli,
  Alobeidli, Pannier, Almazrouei, and Launay}]{penedo2023refinedweb}
Guilherme Penedo, Quentin Malartic, Daniel Hesslow, Ruxandra Cojocaru,
  Alessandro Cappelli, Hamza Alobeidli, Baptiste Pannier, Ebtesam Almazrouei,
  and Julien Launay. 2023.
\newblock The refinedweb dataset for falcon llm: Outperforming curated corpora
  with web data, and web data only.
\newblock \emph{arXiv preprint arXiv:2306.01116}.

\bibitem[{Peng et~al.(2023)Peng, Li, He, Galley, and Gao}]{peng2023instruction}
Baolin Peng, Chunyuan Li, Pengcheng He, Michel Galley, and Jianfeng Gao. 2023.
\newblock Instruction tuning with gpt-4.
\newblock \emph{arXiv preprint arXiv:2304.03277}.

\bibitem[{Rajpurkar et~al.(2016)Rajpurkar, Zhang, Lopyrev, and
  Liang}]{rajpurkar2016squad}
Pranav Rajpurkar, Jian Zhang, Konstantin Lopyrev, and Percy Liang. 2016.
\newblock \href {http://arxiv.org/abs/1606.05250} {Squad: 100,000+ questions
  for machine comprehension of text}.

\bibitem[{Reimers and Gurevych(2019)}]{reimers2019sentencebert}
Nils Reimers and Iryna Gurevych. 2019.
\newblock \href {http://arxiv.org/abs/1908.10084} {Sentence-bert: Sentence
  embeddings using siamese bert-networks}.

\bibitem[{Sanh et~al.(2022)Sanh, Webson, Raffel, Bach, Sutawika, Alyafeai,
  Chaffin, Stiegler, Scao, Raja, Dey, Bari, Xu, Thakker, Sharma, Szczechla,
  Kim, Chhablani, Nayak, Datta, Chang, Jiang, Wang, Manica, Shen, Yong, Pandey,
  Bawden, Wang, Neeraj, Rozen, Sharma, Santilli, Fevry, Fries, Teehan, Bers,
  Biderman, Gao, Wolf, and Rush}]{sanh2022multitask}
Victor Sanh, Albert Webson, Colin Raffel, Stephen~H. Bach, Lintang Sutawika,
  Zaid Alyafeai, Antoine Chaffin, Arnaud Stiegler, Teven~Le Scao, Arun Raja,
  Manan Dey, M~Saiful Bari, Canwen Xu, Urmish Thakker, Shanya~Sharma Sharma,
  Eliza Szczechla, Taewoon Kim, Gunjan Chhablani, Nihal Nayak, Debajyoti Datta,
  Jonathan Chang, Mike Tian-Jian Jiang, Han Wang, Matteo Manica, Sheng Shen,
  Zheng~Xin Yong, Harshit Pandey, Rachel Bawden, Thomas Wang, Trishala Neeraj,
  Jos Rozen, Abheesht Sharma, Andrea Santilli, Thibault Fevry, Jason~Alan
  Fries, Ryan Teehan, Tali Bers, Stella Biderman, Leo Gao, Thomas Wolf, and
  Alexander~M. Rush. 2022.
\newblock \href {http://arxiv.org/abs/2110.08207} {Multitask prompted training
  enables zero-shot task generalization}.

\bibitem[{Scao et~al.(2022)Scao, Fan, Akiki, Pavlick, Ili{\'c}, Hesslow,
  Castagn{\'e}, Luccioni, Yvon, Gall{\'e} et~al.}]{scao2022bloom}
Teven~Le Scao, Angela Fan, Christopher Akiki, Ellie Pavlick, Suzana Ili{\'c},
  Daniel Hesslow, Roman Castagn{\'e}, Alexandra~Sasha Luccioni, Fran{\c{c}}ois
  Yvon, Matthias Gall{\'e}, et~al. 2022.
\newblock Bloom: A 176b-parameter open-access multilingual language model.
\newblock \emph{arXiv preprint arXiv:2211.05100}.

\bibitem[{Scao and Rush(2021)}]{scao2021data}
Teven~Le Scao and Alexander~M. Rush. 2021.
\newblock \href {http://arxiv.org/abs/2103.08493} {How many data points is a
  prompt worth?}

\bibitem[{Schaeffer et~al.(2023)Schaeffer, Miranda, and
  Koyejo}]{schaeffer2023emergent}
Rylan Schaeffer, Brando Miranda, and Sanmi Koyejo. 2023.
\newblock \href {http://arxiv.org/abs/2304.15004} {Are emergent abilities of
  large language models a mirage?}

\bibitem[{Socher et~al.(2013)Socher, Perelygin, Wu, Chuang, Manning, Ng, and
  Potts}]{socher2013recursive}
Richard Socher, Alex Perelygin, Jean Wu, Jason Chuang, Christopher~D Manning,
  Andrew~Y Ng, and Christopher Potts. 2013.
\newblock Recursive deep models for semantic compositionality over a sentiment
  treebank.
\newblock In \emph{Proceedings of the 2013 conference on empirical methods in
  natural language processing}, pages 1631--1642.

\bibitem[{Specia et~al.(2010)Specia, Raj, and Turchi}]{specia2010machine}
Lucia Specia, Dhwaj Raj, and Marco Turchi. 2010.
\newblock Machine translation evaluation versus quality estimation.
\newblock \emph{Machine translation}, 24:39--50.

\bibitem[{Staerman et~al.(2021)Staerman, Mozharovskyi, Colombo,
  Cl{\'e}men{\c{c}}on, and d'Alch{\'e} Buc}]{staerman2021pseudo}
Guillaume Staerman, Pavlo Mozharovskyi, Pierre Colombo, St{\'e}phan
  Cl{\'e}men{\c{c}}on, and Florence d'Alch{\'e} Buc. 2021.
\newblock A pseudo-metric between probability distributions based on
  depth-trimmed regions.
\newblock \emph{arXiv preprint arXiv:2103.12711}.

\bibitem[{Stiennon et~al.(2022)Stiennon, Ouyang, Wu, Ziegler, Lowe, Voss,
  Radford, Amodei, and Christiano}]{stiennon2022learning}
Nisan Stiennon, Long Ouyang, Jeff Wu, Daniel~M. Ziegler, Ryan Lowe, Chelsea
  Voss, Alec Radford, Dario Amodei, and Paul Christiano. 2022.
\newblock \href {http://arxiv.org/abs/2009.01325} {Learning to summarize from
  human feedback}.

\bibitem[{Taori et~al.(2023)Taori, Gulrajani, Zhang, Dubois, Li, Guestrin,
  Liang, and Hashimoto}]{alpaca2023}
Rohan Taori, Ishaan Gulrajani, Tianyi Zhang, Yann Dubois, Xuechen Li, Carlos
  Guestrin, Percy Liang, and Tatsunori~B. Hashimoto. 2023.
\newblock Stanford alpaca: An instruction-following llama model.
\newblock \url{https://github.com/tatsu-lab/stanford_alpaca}.

\bibitem[{Tjong Kim~Sang and
  De~Meulder(2003)}]{tjong-kim-sang-de-meulder-2003-introduction}
Erik~F. Tjong Kim~Sang and Fien De~Meulder. 2003.
\newblock \href {https://www.aclweb.org/anthology/W03-0419} {Introduction to
  the {C}o{NLL}-2003 shared task: Language-independent named entity
  recognition}.
\newblock In \emph{Proceedings of the Seventh Conference on Natural Language
  Learning at {HLT}-{NAACL} 2003}, pages 142--147.

\bibitem[{Touvron et~al.(2023{\natexlab{a}})Touvron, Lavril, Izacard, Martinet,
  Lachaux, Lacroix, Rozi{\`e}re, Goyal, Hambro, Azhar
  et~al.}]{touvron2023llama}
Hugo Touvron, Thibaut Lavril, Gautier Izacard, Xavier Martinet, Marie-Anne
  Lachaux, Timoth{\'e}e Lacroix, Baptiste Rozi{\`e}re, Naman Goyal, Eric
  Hambro, Faisal Azhar, et~al. 2023{\natexlab{a}}.
\newblock Llama: Open and efficient foundation language models.
\newblock \emph{arXiv preprint arXiv:2302.13971}.

\bibitem[{Touvron et~al.(2023{\natexlab{b}})Touvron, Martin, Stone, Albert,
  Almahairi, Babaei, Bashlykov, Batra, Bhargava, Bhosale, Bikel, Blecher,
  Ferrer, Chen, Cucurull, Esiobu, Fernandes, Fu, Fu, Fuller, Gao, Goswami,
  Goyal, Hartshorn, Hosseini, Hou, Inan, Kardas, Kerkez, Khabsa, Kloumann,
  Korenev, Koura, Lachaux, Lavril, Lee, Liskovich, Lu, Mao, Martinet, Mihaylov,
  Mishra, Molybog, Nie, Poulton, Reizenstein, Rungta, Saladi, Schelten, Silva,
  Smith, Subramanian, Tan, Tang, Taylor, Williams, Kuan, Xu, Yan, Zarov, Zhang,
  Fan, Kambadur, Narang, Rodriguez, Stojnic, Edunov, and
  Scialom}]{touvron2023llama2}
Hugo Touvron, Louis Martin, Kevin Stone, Peter Albert, Amjad Almahairi, Yasmine
  Babaei, Nikolay Bashlykov, Soumya Batra, Prajjwal Bhargava, Shruti Bhosale,
  Dan Bikel, Lukas Blecher, Cristian~Canton Ferrer, Moya Chen, Guillem
  Cucurull, David Esiobu, Jude Fernandes, Jeremy Fu, Wenyin Fu, Brian Fuller,
  Cynthia Gao, Vedanuj Goswami, Naman Goyal, Anthony Hartshorn, Saghar
  Hosseini, Rui Hou, Hakan Inan, Marcin Kardas, Viktor Kerkez, Madian Khabsa,
  Isabel Kloumann, Artem Korenev, Punit~Singh Koura, Marie-Anne Lachaux,
  Thibaut Lavril, Jenya Lee, Diana Liskovich, Yinghai Lu, Yuning Mao, Xavier
  Martinet, Todor Mihaylov, Pushkar Mishra, Igor Molybog, Yixin Nie, Andrew
  Poulton, Jeremy Reizenstein, Rashi Rungta, Kalyan Saladi, Alan Schelten, Ruan
  Silva, Eric~Michael Smith, Ranjan Subramanian, Xiaoqing~Ellen Tan, Binh Tang,
  Ross Taylor, Adina Williams, Jian~Xiang Kuan, Puxin Xu, Zheng Yan, Iliyan
  Zarov, Yuchen Zhang, Angela Fan, Melanie Kambadur, Sharan Narang, Aurelien
  Rodriguez, Robert Stojnic, Sergey Edunov, and Thomas Scialom.
  2023{\natexlab{b}}.
\newblock \href {http://arxiv.org/abs/2307.09288} {Llama 2: Open foundation and
  fine-tuned chat models}.

\bibitem[{Wang et~al.(2019)Wang, Singh, Michael, Hill, Levy, and
  Bowman}]{wang2019glue}
Alex Wang, Amanpreet Singh, Julian Michael, Felix Hill, Omer Levy, and
  Samuel~R. Bowman. 2019.
\newblock {GLUE}: A multi-task benchmark and analysis platform for natural
  language understanding.
\newblock In the Proceedings of ICLR.

\bibitem[{Wang et~al.(2023)Wang, Kordi, Mishra, Liu, Smith, Khashabi, and
  Hajishirzi}]{wang2023selfinstruct}
Yizhong Wang, Yeganeh Kordi, Swaroop Mishra, Alisa Liu, Noah~A. Smith, Daniel
  Khashabi, and Hannaneh Hajishirzi. 2023.
\newblock \href {http://arxiv.org/abs/2212.10560} {Self-instruct: Aligning
  language models with self-generated instructions}.

\bibitem[{Wei et~al.(2022{\natexlab{a}})Wei, Bosma, Zhao, Guu, Yu, Lester, Du,
  Dai, and Le}]{wei2022finetuned}
Jason Wei, Maarten Bosma, Vincent~Y. Zhao, Kelvin Guu, Adams~Wei Yu, Brian
  Lester, Nan Du, Andrew~M. Dai, and Quoc~V. Le. 2022{\natexlab{a}}.
\newblock \href {http://arxiv.org/abs/2109.01652} {Finetuned language models
  are zero-shot learners}.

\bibitem[{Wei et~al.(2022{\natexlab{b}})Wei, Tay, Bommasani, Raffel, Zoph,
  Borgeaud, Yogatama, Bosma, Zhou, Metzler, Chi, Hashimoto, Vinyals, Liang,
  Dean, and Fedus}]{wei2022emergent}
Jason Wei, Yi~Tay, Rishi Bommasani, Colin Raffel, Barret Zoph, Sebastian
  Borgeaud, Dani Yogatama, Maarten Bosma, Denny Zhou, Donald Metzler, Ed~H.
  Chi, Tatsunori Hashimoto, Oriol Vinyals, Percy Liang, Jeff Dean, and William
  Fedus. 2022{\natexlab{b}}.
\newblock \href {http://arxiv.org/abs/2206.07682} {Emergent abilities of large
  language models}.

\bibitem[{Williams et~al.(2018)Williams, Nangia, and Bowman}]{N18-1101}
Adina Williams, Nikita Nangia, and Samuel Bowman. 2018.
\newblock \href {http://aclweb.org/anthology/N18-1101} {A broad-coverage
  challenge corpus for sentence understanding through inference}.
\newblock In \emph{Proceedings of the 2018 Conference of the North American
  Chapter of the Association for Computational Linguistics: Human Language
  Technologies, Volume 1 (Long Papers)}, pages 1112--1122. Association for
  Computational Linguistics.

\bibitem[{Xu et~al.(2023{\natexlab{a}})Xu, Sun, Zheng, Geng, Zhao, Feng, Tao,
  and Jiang}]{xu2023wizardlm}
Can Xu, Qingfeng Sun, Kai Zheng, Xiubo Geng, Pu~Zhao, Jiazhan Feng, Chongyang
  Tao, and Daxin Jiang. 2023{\natexlab{a}}.
\newblock \href {http://arxiv.org/abs/2304.12244} {Wizardlm: Empowering large
  language models to follow complex instructions}.

\bibitem[{Xu et~al.(2023{\natexlab{b}})Xu, Wang, Pan, Song, Freitag, Wang, and
  Li}]{xu2023instructscore}
Wenda Xu, Danqing Wang, Liangming Pan, Zhenqiao Song, Markus Freitag,
  William~Yang Wang, and Lei Li. 2023{\natexlab{b}}.
\newblock \href {http://arxiv.org/abs/2305.14282} {Instructscore: Towards
  explainable text generation evaluation with automatic feedback}.

\bibitem[{Zellers et~al.(2019)Zellers, Holtzman, Bisk, Farhadi, and
  Choi}]{zellers2019hellaswag}
Rowan Zellers, Ari Holtzman, Yonatan Bisk, Ali Farhadi, and Yejin Choi. 2019.
\newblock \href {http://arxiv.org/abs/1905.07830} {Hellaswag: Can a machine
  really finish your sentence?}

\bibitem[{Zhang et~al.(2020)Zhang, Kishore, Wu, Weinberger, and
  Artzi}]{zhang2020bertscore}
Tianyi Zhang, Varsha Kishore, Felix Wu, Kilian~Q. Weinberger, and Yoav Artzi.
  2020.
\newblock \href {http://arxiv.org/abs/1904.09675} {Bertscore: Evaluating text
  generation with bert}.

\bibitem[{Zhou et~al.(2023)Zhou, Liu, Xu, Iyer, Sun, Mao, Ma, Efrat, Yu, Yu,
  Zhang, Ghosh, Lewis, Zettlemoyer, and Levy}]{zhou2023lima}
Chunting Zhou, Pengfei Liu, Puxin Xu, Srini Iyer, Jiao Sun, Yuning Mao, Xuezhe
  Ma, Avia Efrat, Ping Yu, Lili Yu, Susan Zhang, Gargi Ghosh, Mike Lewis, Luke
  Zettlemoyer, and Omer Levy. 2023.
\newblock \href {http://arxiv.org/abs/2305.11206} {Lima: Less is more for
  alignment}.

\end{thebibliography}
